# New Approach to translation of Isolated Units in English-Korean Machine Translation

Kim Song Jon, Professor,
Faculty of Foreign Languages and Literature,
**Kim Il Sung** University

An Hae Gum,
Digital Library,
**Kim Il Sung** University

Abstract

It is the most effective way for quick translation of tremendous amount of explosively increasing science and technique information material to develop a practicable machine translation system and introduce it into translation practice. This essay treats problems arising from translation of isolated units on the basis of the practical materials and experiments obtained in the development and introduction of English-Korean machine translation system. In other words, this essay considers establishment of information for isolated units and their Korean equivalents and word order.

We are now faced with the heavy task of developing national science and technology to a new and higher stage by making a revolutionary turn in scientific research work.

Today when the whole country is seething with the spirit of breaking though the extreme points, we are faced with an urgent task to develop the newest domain of science and technique so as to make a contribution to building a powerful state. Requirement for translation which is increasing day by day wants tremendous amount of explosively increasing science and technique information material to be briskly translated and disposed. The most effective way to meet such a requirement is to develop a practicable machine translation system and introduce it into translation practice.   Study of machine translation is profoundly conducted in a practical stage in our country, too.

An isolated unit means a unit which is isolated by commas (in some cases by short lines or by brackets or parentheses ) and takes part in constituting a sentence. To isolated units belong isolated units of noun, attribute and adverbial modifier.

General information of isolated units are set up like this.

·＼Ih＼:The head of an isolated unit

·＼It＼: The tail of an isolated unit

The head information of all kinds of isolated units is placed immediately before the first components of all kinds of isolated units and the tail information placed immediately after the last components of all kinds of isolated units. (2. p. 47)

The detailed properties of isolated units are added to the right side of the head information.

·＼Ih-N＼:The head of an isolated nominal unit

·＼Ih-At＼: The head of an isolated attributive unit

·＼Ih-Ad＼: The head of an isolated adverbial modifying unit

The general information is given to immediately after the detailed isolated units such as nominal units, attributive units and adverbial modifying units. The general tail information is so simply presented because the properties of the types of isolated units are already given together with the head information and so it is unnecessary to present again them together with the tail information.

The following marks and letters are used in this essay.

— (mark): The Korean equivalent before this mark takes a translation order in a left direction.

＋ (mark): The Korean equivalent before this mark takes a translation order in a right direction.

⌢(arc): The section which is denoted by this arc allows a general translation to be conducted and never involves predicates, conjunctions and relatives to be in this section.

Letters or signs, etc. stand for the root (main) part of the Korean equivalent. In a nominal unit of English word, the root part of its Korean equivalent stands for the English word which involves suffixes denoting a singular or a plural number and in an English verbal unit, the root part of its Korean equivalent stands for the English word which indicates its main meaning except for the Korean grammatical suffixes which indicate tense, voice and so forth.

N1: singular noun

N2: plural noun

N: singular or plural noun

P: predicate

S: subject

DET: determinative

L: letter

NUM: number

SB: the beginning of sentence

ReP: relative pronoun

CJS: subordinate conjunction

CJC: coordinative conjunction

G: gerund

Ed: past participle

Ing: present participle,

Prep: preposition

CLB N: noun word whose first letter is a capital one

~N: Adjectives or attributes can be put in front of noun when there is a wave-form mark in its front.

N~: Apposition word may be put in the place of a wave-form mark and its translation is conducted in the right direction.

shoulder number: order mark of words of the same type or the same words

…: There can be any words in this section and the word order of general translation is applied here.

Let us consider translation practice of the above-mentioned isolated units.

1. Isolated Nominal Unit(1, p. 48)

An isolated nominal unit is, in general, an objective unit as an integral unit and it is also often called an appositional unit.

Let us see concrete contextual structures, Korean equivalents and word orders in the examples.

- contextual structure:…CLB N,＼Ih-N＼ DET~N,＼It＼ P DET…

    Korean equivalents and word order: — 《DET+~N》

e.g.: Lastly, only through the collaboration and encouragement by Muriel, \Ih-N \my wife ,\It\ has this project been completed. (마지막으로, 나의 안해 무리엘의 협조와 고무로 하여 이 계획이 완성되였다.)

• contextual structure:…N,\Ih-N\say /namely/ that is, N/L/NUM\It\,…

Korean equivalents and word order: + 《즉/ 다시말하여 N/L/NUM》

e.g.: The solution of Eqs.(1-42) for any current, \Ih-N\ say $I_3$ \It\, using determinants is found by forming a ratio in which the nominator is the determinant of the coefficients of the currents and the numerator is a similar determinant of the coefficients of the unknown current replaced by the right side of the equation. (행렬식들을 리용하여 구한 임의의 전류 즉 $I_3$ 을 위한 식들<1-42>의 해는 분모가 결수들의 행렬식으로 되고 분자는 그 식의 오른쪽에 의하여 교체되는 미지의 전류결수들의 류사한 행렬식으로 되는 비률을 형성함으로써 얻어진다.)

Like this, it is possible to clarify an word order of translation the Korean grammatical suffixes' equivalents to the relevant isolated nominal unit in conformity with the peculiarities of the output language and set forth a word order and Korean equivalents of the connecting means such as adverbs, conjunctions and relatives which connect the basic units and isolated nominal units.

• contextual structure: SB $G^1$…,\Ih-N\or[Adv]$G^2$…\It\, P…

Korean equivalents and word order: + 《즉/다시말하여…[Adv]+ $G^2$》

In this case, if there comes an adverb which realizes emphasis like *even* it is unnecessary to give Korean equivalent to *or*. It is because *or* and *even* both put emphasis into effect and an emphatic meaning is fully felt in the Korean edition without the Korean equivalent to *or* in it.

• contextual structure:…~$N1^1$ ,\Ih-N\$N1^2$ ~\It\,and~$N1^3$ ,\Ih-N\$N1^4$ ~\It\,P…

Korean equivalent and word order: …~$N1^1$ 인+ 《$N1^2$ ~》 와/과 + ~$N1^3$ 인 + 《$N1^4$ ~》

Here, if 《인》, a Korean appositional grammatical suffix, is added to the Korean equivalent to the unit in front of an isolated unit, the compiling of the whole Korean equivalents becomes more natural.

• ·contextual structure :~$N^1$, \Ih-N\or ~$N^2$[L] \It\, P…

Korean equivalents and word order: + 《즉 ~N²[L]》

2. Isolated Attributive Unit(1, p. 52)

Isolated attributive units are treated as attributes whose translation word order is in the left direction by setting forth the head and the tail information as in other isolated units. In other words, the word orders of the Korean attributive equivalents are decided so that such equivalents are be placed in front of the attributive units in clausal and phrasal attributes which are isolated attributive units.

Let us see translation of isolated attributive units.

• contextual structure :SB [Adv] S,\Ih-At\ with DET¹ ~N¹ ⌒¹ and DET² ~N²⌒²\It\,P…

Korean equivalents and word order: — 《⌒¹에/에는 DET¹ ~N¹이 있고 ⌒²에/에는 DET² ~N²이 있는》

례: Obviously a potentiometer, \Ih-At\ with its terminals at each end of the resistant element and a third terminal attached to the slider \It\, can be used as a thermostat if one of the resistance-element terminals if ignored. (틀림없이 <u>저항요소의 매 끝에는 단자들이 있고 세번째 단자가 미끄럼편에 붙어있는</u> 전위차계는 저항요소단자들가운데서 하나가 무시된다면 온도조절계로 리용될수 없다.)

Like this, an isolated unit consists of homogeneous units before and after the coordinative conjunction which are composed of the same elements in the same structure and in this case, a general translation word order is applied within the homogeneous units and such word order is applied between homogeneous units in the right direction.

• contextual structure:…N⌒¹ ,\Ih-At\ ReP P¹ ⌒² \It\,P² …

Korean equivalent and word order: — 《⌒² P¹ 는》

e.g.: In simple dc circuits the source of this energy, \Ih-At\which must be supplied in order to maintain the current\It\, is often a chemical battery.(단일 직류회로에서 <u>전류를 유지하기 위하여 공급되여야 하는</u> 이 에네르기의 원천은 종종 화학전지이다.)

• contextual structure:…N¹ ,\Ih-At\Ed Prep ~N² \It\, and S P…

Korean equivalents and word order: — 《~N² Prep Ed ㄴ/는》

례: Those goods and services should be classified according to an international classification, \Ih-At\known as the Nice Classification\It\, and the International Bureau has general responsibility for the consistent application of the Classification. (그 상품 및 봉사들은 <u>나이스 클래씨휘케이슌으로 알려진</u> 국제분류에 따라 분류되여야 하며 국제뷰로는 그 분류의 일관한 적용에 대한 책임을 진다.)

Like this, it is possible to set forth the combination structure of isolated attributive units, formalization of their front and back contextual conditions, their necessary Korean equivalents and word orders..

3. Isolated Adverbial Modifying Unit (1, p. 53)

An isolated adverbial modifying unit begins with subordinate conjunction or participle or preposition, etc and functions as adverbial modifier. The Korean equivalent of this isolated unit is added to that of the unit in front of it by giving the head and the tail information in the isolated adverbial modifying unit, too, as in the above case.

It is possible to formalize an isolated adverbial modifying unit and accordingly realize translation like this.

• contextual structure :…, \Ih-Ad\if　[negative Adv]　Adv Ed\It\, …

　Korean equivalents and word order: — 《Adv Ed [지 않는다 해도] /ㄴ다 해도》

Some English conjunctions or other English words are special in their Korean equivalents. In this case, relevant English words can be presented in the contextual structure as they are.

e.g.: It is always assumed, \Ih-Ad\if not explicitly indicated\It\, that the longer line represents the higher, or positive, terminal of the internal *emf*. (<u>단정적으로 말할수는 없으나</u> 더 긴 도선이 내부기전력의 더 높은 단자 즉 양전기단자라고들 항상 생각한다.)

Like this, it is possible to set forth the combination structure of isolated adverbial modifying units, formalization of their front and back contextual conditions, their necessary Korean equivalents and word order.

• ·contextual structure:…, \Ih-Ad\Ing⌒\It\,…

　Korean equivalents and word order: — 《⌒Ing》

e.g.: Now, in order to determine the equivalent resistance for practical resistors, we define $R_1$ , \Ih-Ad\ using Ohm's law \It\, as…(이제는 실제저항기에 맞먹는 저항을 결정하기 위하여 우리는 <u>옴의 법칙에 따라</u> 저항 $R_1$ 을 …라고 규정한다.)

Like this, we've considered the translation of isolated units in EKMT.

We will make a contribution to the progress of science and technique for building a socialist powerful state by further completing EKMT system.

References:

1.Kim Song Jon(2012), Comprehensive Corpus and Patterns for English-Korean Machine Translation. Publishing House of Science and Technique, Pyongyang, DPR of Korea

2.Doronthy Kenny & Andy Way (2007). Teaching Machine Translation and Translation Technology. Dublin city University, Dublin, Ireland

3. Deng, Y. and Byrne, W. (2005). HMM word and phrase alignment for statistical machine translation. In *HLT-EMNLP-05*

4. Chiang, D. (2005). A hierarchical phrase-based model for statistical machine translation. In *ACL-05*, Ann Arbor, MI, pp.263-270. ACL

5. Koehn, P., Och, F. J., and Marcu, D. (2003). Statistical phrase-based translation. In *HLT-NAACL-03*, pp.48-54